\documentclass[journal]{IEEEtran}

\usepackage[utf8]{inputenc}
\usepackage{cite}

\usepackage[caption=false, font=footnotesize]{subfig}
\usepackage{placeins}
\usepackage{pgfplots}
\usepackage{amsmath}
\usepackage{amssymb}
\usepackage{amsthm}
\usepackage{algorithm}
\usepackage{algorithmic}

\usepackage{tikz}

\usepackage{todonotes}

\usepackage{ifthen}

\newboolean{PrePrintWatermark}
\setboolean{PrePrintWatermark}{false}

\newboolean{PrintVersion}
\setboolean{PrintVersion}{true}

\ifCLASSINFOpdf
\else
\fi

\newcommand\copyrighttext{%
	\footnotesize 
	SUBMITTED TO REVIEW AND POSSIBLE PUBLICATION. COPYRIGHT WILL BE TRANSFERRED WITHOUT NOTICE. 
	Personal use of this material is permitted.
	Permission must be obtained for all other uses, in any current or future media, including reprinting/republishing this material for advertising or promotional purposes, creating new collective works, for resale or redistribution to servers or lists, or reuse of any copyrighted component of this work in other works.}%
\newcommand\copyrightnotice{%
	\begin{tikzpicture}[remember picture,overlay]%
	\node[anchor=south,yshift=-50pt] at (current page.north) {\fbox{\parbox{\dimexpr\textwidth-2cm}{\copyrighttext}}};%
	\end{tikzpicture}%
	\vspace{-10pt}%
}

\begin{document}

\title{Motion Planning for Connected Automated Vehicles at Occluded Intersections With Infrastructure Sensors}

\author{Johannes~Müller,~\IEEEmembership{Student Member,~IEEE,}
        Jan~Strohbeck,~\IEEEmembership{Student Member,~IEEE,}
        Martin~Herrmann,~\IEEEmembership{Student Member,~IEEE,}
        and~Michael~Buchholz
\thanks{Part of this work was financially supported by the Federal Ministry of Economic Affairs and Energy of Germany within the program "Highly and Fully Automated Driving in Demanding Driving Situations" (project MEC-View, grant number 19A16010I). Part of this work has been conducted as part of ICT4CART project which has received funding from the European Union’s Horizon 2020 research \& innovation program under grant agreement No. 768953. Content reflects only the authors’ view and European Commission is not responsible for any use that may be made of the information it contains. }
\thanks{J. Müller, J. Strohbeck, M. Herrmann, and M. Buchholz are with the Institute of Measurement, Control and Microtechnology, Ulm University, Ulm, 89134 Germany e-mail: johannes-christian.mueller@uni-ulm.de,\{first name.family name@uni-ulm.de\}.}
\thanks{Manuscript received November 23, 2020; revised August 05, 2021.}}

\maketitle
\copyrightnotice%

\begin{abstract}
Motion planning at urban intersections that accounts for the situation context, handles occlusions, and deals with measurement and prediction uncertainty is a major challenge on the way to urban automated driving.
In this work, we address this challenge with a sampling-based optimization approach.
For this, we formulate an optimal control problem that optimizes for low risk and high passenger comfort.
The risk is calculated on the basis of the perception information and the respective uncertainty using a risk model.
The risk model combines set-based methods and probabilistic approaches.
Thus, the approach provides safety guarantees in a probabilistic sense, while for a vanishing risk, the formal safety guarantees of the set-based methods are inherited.
By exploring all available behavior options, our approach solves decision making and longitudinal trajectory planning in one step. 
The available behavior options are provided by a formal representation of the situation context, which is also used to reduce calculation efforts.
Occlusions are resolved using the external perception of infrastructure-mounted sensors.
Yet, instead of merging external and ego perception with track-to-track fusion, the information is used in parallel.
The motion planning scheme is validated through real-world experiments.
\end{abstract}


\IEEEpeerreviewmaketitle

\section{INTRODUCTION}

Urban automated driving has been a field of intense study for the last decades, yet robust automated driving in urban areas has not been achieved so far \cite{Yurtsever2020}.
Intersections are a particular challenge.
Accident statistics \cite{Simon2009} suggest that even among human drivers, safely navigating through junctions is challenging:
other road users need to be perceived, tracked and predicted over long time horizons, which inherently comes with high uncertainties.
Under these uncertainties, possible maneuver options have to be explored and the most suitable option has to be decided for, which includes an appropriate timing.
The decisions of the human driver depend strongly on the situation context.
For a technical system, all this has to be carried out in real time with the utmost robustness and reliability, which is still an unsolved challenge \cite{Yurtsever2020,Eskandarian2019,Hussain2019}.
In addition, many junctions are heavily occluded, which poses an additional open issue for motion planning \cite{Hussain2019}.

\begin{figure}[tbb]
	\centering
	\includegraphics{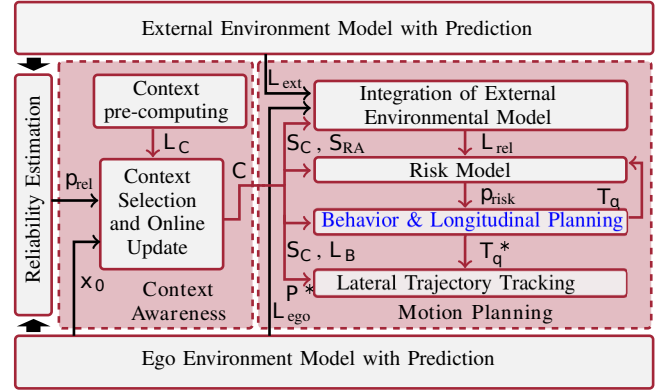}	
	\caption{Architecture overview on the  proposed planning approach.}
	\label{fig:Overview}
\end{figure}

In this work, we address these challenges with a new motion planning framework with the architecture shown in Figure~\ref{fig:Overview}.
The core of this framework is a combined behavior and longitudinal trajectory planning, which is based on our previous work \cite{MuellerITSC2019}.
The combined planning is formulated as an Optimal Control Problem (OCP) over the longitudinal states position $s_\shortparallel$, velocity $v_\shortparallel$, and acceleration $a_\shortparallel$, optimizing for risk and passenger comfort.
To solve the OCP, first, a formal representation of the situation context is computed, which comprises the set of active constraints $\mathcal{S}_C$, a set of regularizing assumptions $S_\text{RA}$ (e.g., the rational driver assumption), and a list of available behavior options $\mathcal{L}_\mathcal{B}$. 
The behavior options serve as templates from which trajectory candidates $\mathcal{T}_\shortparallel$ are generated using an analytical partial solution of the OCP. 
Hence, our approach can roughly be categorized as being sample-based.
Since calculating the analytical solution is very fast, the planning scheme can explore several hundred behavior options across different behavior types in real time.
The trajectory candidates are checked to comply with the constraints $\mathcal{S}_C$, their residual risk $p_\text{risk}(\mathcal{T}_\shortparallel)$ is calculated using our risk model, their overall costs are determined, and the overall optimal and feasible trajectory $\mathcal{T}_\shortparallel^\ast$ is chosen,
where the existence of a feasible $\mathcal{T}_\shortparallel^\ast$ is guaranteed by a fail-safe strategy.
$\mathcal{T}_\shortparallel^\ast$ then is stabilized along the selected reference path by the lateral trajectory tracking.
The advantage of the proposed architecture compared to classical ones \cite{Paden2016} can be illustrated at merging into a junction.
A classical behavior layer would have to specify the traffic gap to which the automated vehicle should merge. 
Thus, it would need, e.g., a state in a state machine for each decision.
In contrast, our approach would consider all available gaps and would choose the trajectory with the globally optimal cost.

To the best of our knowledge, we are the first to propose a risk model that combines set-based and probabilistic methods.	
Based on the perception reliability $p_\text{rel}$ calculated by our method from \cite{MuellerIV2019} and the environment model with uncertainties, the risk model computes the probability $p_\text{risk}(\mathcal{T}_\shortparallel)$ that the candidate trajectory leaves its \emph{invariably safe set} (ISS) as defined in \cite{Pek2017}. 
The ISS comprises states, which for all times $t$ up to the prediction horizon $T_\text{pred}$ are the starting point of at least one feasible trajectory that again ends in a ISS.
From this concept, we inherit formal safety guarantees for the special case of a vanishing residual risk, while these guarantees are otherwise softened to guarantees in a probabilistic sense.

In contrast to most partially observable Markov decision processes (POMDPs) based methods (e.g. \cite{Hubmann2018a}), we consider the interaction between
other road users and the ego vehicle as a constraint and minimize the risk of violating this constraint.
Thus, we argue that for our problem formulation, the prediction of other road users is only loosely coupled with the motion planning.
As a result, we expect only an insignificant reduction in performance when solving the planning and prediction problem independently.
Occlusions are resolved using an external environment model.
For a connected automated vehicle (CAV) with vehicle-to-anything (V2X) communication capabilities, such a model can be received from different sources, e.g., from a road-side unit or a multi-access edge computing (MEC) server. 	 
Furthermore, they can be computed from multiple sources, e.g., from shared information of connected road users or from infrastructure sensors.
For the evaluation in this work, we use an external environment model that is provided via a low-latency mobile network by a MEC server based on infrastructure sensors \cite{Herrmann2019}.
In the vehicle, instead of classical T2T fusion with the internal environment model, we propose to use the external environment model in parallel.
Thus, self-fulfilling prophecies can be avoided that might occur if the CAV's environment model is passed back to the environment model service.

The proposed framework extends our previous work\cite{MuellerITSC2019} in several ways. 
First, context awareness is added to generalize the approach to a broad variety of situations, accounting for additional constraints. Particularly, the lateral dynamics are considered. Second, the risk model is formalized and generalized so that it is possible to use external information, which might not be fully reliable. Furthermore, the generalized risk model allows to consider more than two relevant road users. Thus, it describes many situations more accurately. Third, the new planning scheme is extended to use target sequences instead of target states. Thus, it is possible to drive more complex motion patterns and, thus, fulfill more complex motion tasks. For example, driving through narrow curves requires a step-like velocity profile, which is hardly achievable with a single target state. Fourth, a lateral trajectory tracking is added, which stabilizes the CAV and, thus, significantly reduces the requirements to the subsidiary control of the CAV.

\subsection{Related Work} \label{sec:RelatedWork}
In classical approaches, e.g. \cite{Puphal2018,Voelz2019,Orzechowski2020}, motion planning is divided into decision making and trajectory planning, and the two subproblems are solved individually \cite{Paden2016}.
Yet, this comes with the drawback that the trajectory planning might not be able to safely execute the commands of the decision layer.

More recent approaches based on POMDPs use reinforcement learning \cite{Gritschneder2016,Meghjani2019} or related methods \cite{Hubmann2018a} for decision making (e.g. \cite{Meghjani2019}), or to solve the motion planning problem  holistically.
However, solving POMDPs is PSPACE-complete in general\cite{Schwarting2018}.
Hence, these approaches usually suffer from high computational burden.

With the rise of deep learning, end-to-end (E2E) learning approaches, e.g. \cite{Kendall2019}, got into focus of current research and already show promising results.
However, following the arguments of \cite{Yurtsever2020}, modular systems are favorable over E2E approaches, as they feature better interpretability.
Moreover, safety measures are harder to apply to E2E systems \cite{Yurtsever2020}, lots of training data is needed for informed decision making \cite{Hussain2019}, and bad generalization was reported for E2E systems \cite{Yurtsever2020}.

Set-based methods, e.g. \cite{Pek2017,Pek2018,Ge2019}, focus on the safety aspect of automated driving and come with the appealing benefit that they feature formal safety guarantees.
Formal safety guarantees are a key goal and still a major challenge on the way to automated driving \cite{Schwarting2018}.
In fact, our proposed risk model builds upon the set-based methods and thus achieves the same formal guarantees if the residual risk vanishes.
However, the set-based methods are not easy to handle, as they usually require a large set of situation-dependent rules and regularizing assumptions \cite{Pek2018}.
Particularly, most sensors and perception modules are modeled and characterized probabilistically. 
Therefore, Ge et al. \cite{Ge2019} themselves admitted that further research towards a probabilistic extension of the set-based methods, as the one we propose in this work, is necessary.
Other works, like \cite{Ward2018,Kumar2018}, have also addressed risk minimized planning, but none of them includes real-world experiments nor features formal guarantees as far-reaching as those of the set-based methods.

Communication-based approaches, in turn, show very promising results, in particular, as other road users' intents and future trajectories do not need to be inferred, but are communicated.
However, these approaches assume all vehicles to be interconnected, which is not the case for current mixed traffic.
We refer to the surveys \cite{RiosTorres2017} and \cite{Pereira2017} for details.

Other approaches, e.g. \cite{Guenther2016,Ambrosin2019}, as in this work, use collective perception, e.g., to resolve occlusions. To be scalable with respect to the data load on the network, these approaches share preprocessed object lists rather than raw data. These object lists then are merged using a track-to-track (T2T) approach \cite{Radtke2019}. However, T2T fusion suffers from the problem that the incoming data has already been filtered and is therefore no longer statistically independent, as most Bayesian approaches require. 
Hence, optimal T2T fusion requires effortful correction mechanisms \cite{Radtke2019}. In contrast, typically used sub-optimal fusion methods overestimate the covariance of the states of the objects.

According to \cite{Hussain2019}, context-aware motion planning schemes promise great potential for key challenges like coping with unpredictable environments, but have not yet received enough attention in research.
In \cite{Casas2018}, static situation context was used to predict the intention of other road users, while in \cite{Meghjani2019}, static context was additionally used for decision making.
However, none of these works used static and dynamic situation context for the complete motion planning task, as proposed in this paper.

\subsection{Contributions}\label{sec:Contributions}
The contribution of this work can be summarized as follows:
\begin{itemize}
	\item We propose a formalized situation context representation that allows the CAV to adapt to different situations. Furthermore, it allows to reduce the calculation efforts.
	\item Our approach optimizes for safety and passenger comfort over all available behavior options across behavior types. That is, decision making and longitudinal trajectory planning are solved in one step.	
	\item The approach in general provides safety guarantees in a probabilistic sense, while in case of a vanishing residual risk, the formal safety guarantees from \cite{Pek2017} are inherited.
	\item The proposed risk model can account for the additional challenge that external information might not be reliable.	
	\item The approach is validated in the challenging real-world experiment of merging into an occluded yield junction with external perception.
\end{itemize}

\subsection{Structure of this Work}\label{sec:Structure}
The rest of the paper is structured as follows: Section \ref{sec:ApplicationScenario} presents the application scenario used for experimental validation, which also serves to illustrate the motion planning scheme.
Our concept of context awareness is presented in Section \ref{sec:contextAwareness}, and the motion planning scheme is discussed in Section \ref{sec:Planning}.
Finally, the proposed approach is validated through real world experiments in Section \ref{sec:ExperimentalValidation}, and the article closes with some conclusions and an outlook on future work in Section~\ref{sec:Conclusion}.

\section{APPLICATION SCENARIO AND EXPERIMENTAL SETUP} \label{sec:ApplicationScenario}
\begin{figure}[tbp]
	\centering
	\includegraphics[width=0.30\textwidth]{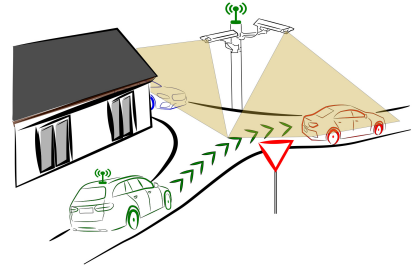}
	\caption[Application Scenario]{Application Scenario.}
	\label{fig:ApplicationScenario}
\end{figure}

The motion planning scheme presented in this work generalizes to many situations, yet needs to be parametrized accordingly.
To bridge the gap between the general but abstract motion planning scheme and its concise application, a challenging traffic situation is presented as an exemplary application scenario.
In this scenario, the CAV uses external information from infrastructure-based sensors to dynamically merge into an occluded yield junction, see Figure \ref{fig:ApplicationScenario}.
For evaluation, we developed an overall system architecture, which was implemented at a pilot junction in Ulm-Lehr \cite{Buchholz2020}.

At this junction, the CAV approaches a suburban yield junction occluded by a house close to the road, trying to dynamically merge into the junction.
To do so, the CAV requires external information to resolve the occlusions.
This information is provided by infrastructure-mounted sensors that detect oncoming traffic at the junction and send these detections to a MEC server via a mobile network.
On the MEC server, the detections of multiple sensors are merged and tracked to build an environment model of the junction \cite{Herrmann2019}.
Additionally, to compensate for latency and to support predictive planning, a prediction of the environment model is performed on the MEC server.
In the simplest case, rough predictions can be obtained using a Kalman filter.
Such simple predictions are used in this work to show robustness of the planning scheme.
However, more recent prediction approaches, e.g. \cite{Strohbeck2020}, can be used to get predictions with higher accuracy and thus allow for even higher motion performance.
Finally, the environment model, represented by a list of tracked and predicted objects $\mathcal{L}_{\text{ext}}$, is sent to the CAV via the mobile network.
Based on the list of external objects $\mathcal{L}_{\text{ext}}$ and the list of objects $\mathcal{L}_{\text{ego}}$ that the CAV perceives with its ego perception, the CAV plans its motion such that it can dynamically merge into a traffic gap at the junction.

The CAV used for experimental validation is the automated vehicle described in \cite{Buchholz2020}.
It is equipped with a multi-sensor setup consisting of camera, lidar, and radar sensors.
From the sensors' information, detection lists are generated that are then used to track the objects in the scene.
Furhermore, the CAV localizes itself with a high-precision RTK system, from which the ego state estimation is derived.
For experimental validation, the CAV is operated around the pilot junction in Ulm-Lehr \cite{Buchholz2020}, which is equipped with camera and lidar infrastructure sensors.
These external sensors cover a field of view  (FOV) of about $85~\text{m}$ from the junction, which corresponds to a maximum prediction horizon of $T_\text{pred}=10~\text{s}$ at the speed limit of $v_\text{sl} = 30~\text{km/h}\approx 8.33~\frac{\text{m}}{\text{s}}$.
For communication, the pilot junction is covered by a mobile network that is optimized to transmit the environment model with low latency of about $10-20$~ms \cite{Buchholz2020}.
For safety considerations, the lateral acceleration is constrained conservatively to a maximum lateral acceleration of $a_{\perp,\text{max}}=1.45~\tfrac{\text{m}}{\text{s}^2}$.
The pilot junction contains a challenging narrow S curve shape when driving from west to east. It shows a minimum curve radius of $R \approx 12.5~\text{m}$.
The longitudinal acceleration of the CAV is limited to $a_\shortparallel \in [-4~\tfrac{\text{m}}{\text{s}^2},\,2~\tfrac{\text{m}}{\text{s}^2}]$.

\section{CONTEXT AWARENESS}\label{sec:contextAwareness}

In this section, the concept of context awareness is presented first in general and then illustrated for our exemplary application scenario.
The key motivation for context awareness is to explicitly use situation context to first parametrize the motion planning problem to the current situation and then to simplify  it as much as possible.
Thus, the CAV can adapt its behavior to different traffic scenarios and, for most situations, the computational load can be reduced.
Moreover, with the situation context, situation specific fail-safe strategies can be determined and the CAV's behavior can be tailored such that safety can be formally guaranteed for many situations.
Situation context, on the one hand, is based on static information like the traffic rules that can be precomputed from the map.
On the other hand, the situation context is updated online through perception information.
For example, if another road user is detected ahead of the CAV, the static speed limit $v_\text{max}(s_\shortparallel)$ is overwritten by a dynamic speed limit profile $v_\text{max}(s_\shortparallel,t)$, where $t$ denotes the time and $s_\shortparallel$ the 1D position. This prevents the CAV from colliding with that road user.

\subsection{General Concept} \label{sec:GeneralConcept}
For this work, situation context is represented as
\begin{equation}
\mathcal{C}_{[s_j,s_{j+1})} = (\mathcal{S}_C,\mathcal{S}_{\text{RA}},\mathcal{L}_\mathcal{B}) \, ,
\end{equation}
with the following four basic parts: an interval $[s_j,s_{j+1})$ stating for which part of the route the respective context is valid, a set of constraints $\mathcal{S}_C$ that apply to the CAV as well as to other road users, a set of regularizing assumptions $\mathcal{S}_{\text{RA}}$ used to parametrize the risk model as well as the planning, and a list of behavior options $\mathcal{L}_\mathcal{B}$.
Intuitively, within one situation, the constraints and regularizing assumptions stay the same.
However, $\mathcal{S}_C$ and $\mathcal{S}_{\text{RA}}$ are constant only on segments of the route and generally depend on $s_\shortparallel$.
Hence, to make $\mathcal{S}_C$ and $\mathcal{S}_{\text{RA}}$ invariant to $s_\shortparallel$, the route is split into segments $[s_j,s_{j+1}]$, on which the situation context is defined.

The set of constraints $\mathcal{S}_C$ comprises mathematical statements requiring a variable to be included in or excluded from a specific set.
Such set can be described, e.g., by a function, which results in equality constraints, by upper and lower bounds, which yields inequality constraints, or by a polygon.
For example, $\mathcal{S}_C$ contains the traffic rules, i.e., the legal speed limit $v_\text{sl}(s_\shortparallel)$, mandatory stops, yield lines, the safety distances $s^{-}(t), \, s^{+}(t)$ to other road users behind and in front of the CAV, etc.
Furthermore, priority lanes are defined by polygons.
The CAV has to give way to road users on such priority lanes and thus has to include the corresponding objects in $\mathcal{L}_\text{rel}$. 
Moreover, $\mathcal{S}_C$ contains the acceleration bounds $a_\text{min}, \, a_\text{max}$, and the speed limit profile
\begin{equation}
\label{eq:speed_limit_profile}
v_{\text{max}}(s_\shortparallel) = \min \left\{ v_{\text{sl}}(s_\shortparallel), \, \sqrt{\frac{|| a_{\perp,\text{max}}||}{\kappa_\text{ref}(s_\shortparallel)}} \right\} \, ,
\end{equation}
where $\kappa_\text{ref}(s_\shortparallel)$ is the curvature of the road, which can be calculated, e.g., with the method from \cite{Gritschneder2018}.

The set of regularizing assumptions $\mathcal{S}_{\text{RA}}$ is of the same form as $\mathcal{S}_C$. 
However, they describe assumptions, which ensure the planning problem is well-posed and furthermore help our algorithm to find good solutions quickly.
For example, $\mathcal{S}_{\text{RA}}$ includes the rational driver assumption, which requires the other road users to follow the traffic rules.
Hence, road users having to give way to the CAV can be excluded from $\mathcal{L}_\text{rel}$.
As a further example, $\mathcal{S}_{\text{RA}}$ includes the assumption on the perception, which is specified in terms of polygons describing the FOVs and the estimated perception reliability $p_\text{rel}$.
Finally, $\mathcal{S}_{\text{RA}}$ contains some heuristics to reduce the calculation efforts of the planning.
For example, it is assumed that the passengers want to drive through curves with constant velocity.
This allows to approximate $v_{\text{max}}(s_\shortparallel)$ by a segment-wise constant function, where $v_{\text{max}}(s_\shortparallel)$ is either $v_{\text{sl}}(s_\shortparallel)$, or the velocity at the minimum curve radius.
Moreover, heuristics allow to prioritize the available behavior options. This allows the algorithm to return prematurely, if a favored solution is found.

Finally, the list of behavior options $\mathcal{L}_\mathcal{B}$ defines the sampling space of the longitudinal planner, where each behavior option $\mathcal{B}$ can be used to generate a trajectory candidate $\mathcal{T}_\shortparallel$.
A behavior option can be formalized as
\begin{equation}
\label{eq:BehaviorOption}
\mathcal{B}: ( \mathcal{P}, \, \langle(\boldsymbol{x}_f,t_f)\rangle_{i}, \, I ) \, ,
\end{equation}
where $\mathcal{P}$ refers to the path the CAV is intended to follow,
$\langle(\boldsymbol{x}_f,t_f)\rangle_{i}, \, i=1, \dots , N_t$
is a target sequence of $N_t$ target states $\boldsymbol{x}_{f,i}=[s_{\shortparallel,v,i},s_{\shortparallel,f,i},a_{\shortparallel,f,i}]$ and respective final times $t_{f,i}$, and $I$ is the importance of the respective behavior option.

As required by the lateral trajectory tracking, the reference path $\mathcal{P}$ is represented by its curvature $\kappa_\text{ref}(s_\shortparallel)$.
It describes a 2D curve on the map $\mathcal{M}$.
Many behavior options share the same $\mathcal{P}$, which often simply is the curvature of the lane at the position $s_\shortparallel$. However, for overtaking or obstacle avoidance, different paths $\mathcal{P}$ are generated, e.g. with the method from \cite{Thompson2005}.

With the target sequence $\langle(\boldsymbol{x}_f,t_f)\rangle_{i}$, the timing of the behavior option is specified.
Particularly, the target sequence provides key states the CAV has to pass on its trajectory at a specific time $t_{f,i}$. 
As opposed to using a single target state $\boldsymbol{x}_f(t_f)$ at the final time $t_f$, the additional targets $(x_f,t_f)$ add further degrees of freedom to the planning.
Hence, more complex motion patterns are possible.
However, the additional degrees of freedom also increase the computational cost.
Therefore, $N_t$ is kept short, where we use $N_t \leq 3$ in practice.
To ensure consistency between the situation contexts, each behavior option is required to have a target state at the boundary between the contexts, which fulfills the constraints of the updated current context and its pre-computed successor.

Finally, for known situations, the heuristics in $\mathcal{S}_\text{RA}$ are used to assign an importance weight $I$ to the behavior options.
Thus, the planning algorithm can explore the available behavior options in the order of their importance $I$ and return prematurely, if the remaining behavior options are less favored.
Note that the importance weighting is different from predefined behavior in conjunction with finite state machines as used in classical approaches \cite{Paden2016}.
For unknown situations, in our approach, all behavior options would be weighted equally due to the lack of an adequate classification rule, i.e. heuristics.
Conversely, the lack of an adequate classification rule would cause a finite state machine to end up in an undefined state.
Thus, undefined behavior would result in those approaches.
\subsection{Processing of the Situation Context}
The processing of the situation context comprises two stages: the pre-computing and the online selection and update of the currently active situation context.
During pre-computing, the road geometry along the route and the traffic rules are extracted from the map $\mathcal{M}$.
From this, $\mathcal{S}_C$ and $\mathcal{S}_\text{RA}$ are derived, both in dependence of the position $s_\shortparallel$.
Then, the route is split into segments $[s_j,s_{j+1}]$ to which the same $\mathcal{S}_C$ and $\mathcal{S}_\text{RA}$ apply.
Lastly, some behavior options are pre-computed.
This primarily includes the behavior options corresponding to the fail-safe strategy.
In urban areas, this usually is constant braking to standstill with $a_\text{min}$.
It is further assumed that the CAV starts at $s_\shortparallel(0)=s_j$ and drives to the end of the situation context. 
Finally, the consecutive situation contexts are stacked into the list of situation contexts $\mathcal{L}_\mathcal{C}$.

During online selection and update, first, the currently active situation context is selected.
For that, the current ego state is projected to the road, which yields $s_\shortparallel(0)$. Then, the context is chosen, for which $s_j \leq s_\shortparallel(0) < s_{j+1}$ holds.
Next, the respective context is updated.
First, current measurements are used to update $\mathcal{S}_C$ and $\mathcal{S}_\text{RA}$.
For example, if a vehicle is detected ahead of the CAV's lane, the static speed limit profile $v_\text{max}(s_\shortparallel)$ is restricted to the dynamic speed limit profile 
\begin{equation}
\label{eq:dyn_speed_limit_profile}
v_\text{max}(s_\shortparallel,t) = \begin{cases}
v_\text{max}(t) \, , \quad s_\shortparallel < s_\text{pred}(t) - s^{+}(t) \, , \\
v_\text{pred}(t) \, , \quad \text{otherwise} \, ,
\end{cases}
\end{equation}
where $s_\text{pred}$, $v_\text{pred}$ are the vehicle's predicted position and velocity, respectively.
Thus, no valid trajectory candidate enters the \emph{reachable set}\cite{Pek2017} of the vehicle ahead.
This way, it is ensured that the CAV does not collide with this vehicle. 
As a further example, the rational driver assumption is dropped for road users in clear violation of the traffic rules.

Second, the pre-computed behavior options are checked to comply with the updated $\mathcal{S}_C$, $\mathcal{S}_\text{RA}$ and invalid behavior options are dropped.
Furthermore, additional behavior options are generated.
For example, behavior options with varying reference paths $\mathcal{P}$ are generated for obstacle avoidance or overtaking, if an object ahead of the CAV occurs.
Moreover, the behavior options are updated to the current situation.
For this, targets $(\boldsymbol{x}_f,t_f)$ that the CAV has already passed are dropped and the final times of the target sequences are adjusted such that $s_\shortparallel(0)$ is on the corresponding trajectory.

At this stage, the set-based methods \cite{Pek2017,Pek2018,Ge2019} are employed to obtain safety guarantees.
First, the \emph{reachable set} of all perceived objects is calculated.
From this, the sets $\hat{\mathcal{S}}_\text{ISS}$, which are known to be subsets of the CAV's ISS, are calculated as described in \cite{Pek2017}, using the pre-computed fail-safe strategy, $\mathcal{S}_C$, and $\mathcal{S}_\text{RA}$.
These subsets $\hat{\mathcal{S}}_\text{ISS}$ are used as an approximation to the CAV's ISS, since determining the ISS exactly is computationally demanding.
For safety reasons, all valid behavior options are required to end in a set $\hat{\mathcal{S}}_\text{ISS}$.
The state at which a trajectory enters this $\hat{\mathcal{S}}_\text{ISS}$ is termed \emph{point of guaranteed arrival} (PGA) \cite{Pek2017}. 
If a behavior option is fully included within a set $\hat{\mathcal{S}}_\text{ISS}$, safety is formally guaranteed \cite{Pek2017}, and $p_\text{risk}=0$.
Otherwise, its target sequence is modified such that the final target state $\boldsymbol{x}_{f,N_t}$ becomes the PGA of the behavior option.
If a behavior option connects two sets $\hat{\mathcal{S}}_\text{ISS}$, the state on the corresponding trajectory at which the CAV leaves the first $\hat{\mathcal{S}}_\text{ISS}$ is termed \emph{point of no return} (PNR) \cite{Pek2017}.
Behavior options which contain a PNR are modified such that one of the target states becomes the PNR.
The PNRs and PGAs are required by the risk model to compute the residual risk (see Section~\ref{sec:RiskModel}).

\subsection{Situation Context for the Application Scenario} \label{sec:COntextApplicationScenario}
The general situation context concept is now applied to the application scenario.
First, the path $\mathcal{P}$ is determined from $\mathcal{M}$. 
Then, the acceleration bounds are loaded and the speed limit profile $v_\text{max}(s_\shortparallel)$ is computed with \eqref{eq:speed_limit_profile}.
The speed limit before and after the junction is $v_\text{max}(s_\shortparallel) \approx 8.33~\frac{\text{m}}{\text{s}}$, while due to \eqref{eq:speed_limit_profile}, the CAV has to slow down to about $4~\frac{\text{m}}{\text{s}}$ for the narrow S-shaped curve.
The traffic rules obtained from $\mathcal{M}$ require the CAV to yield at $s_\text{stop}$ and to give way to vehicles approaching from left.
The fail-safe strategy for the application scenario is braking with $a_\text{min}$.
Accordingly, the PNRs are the states from which stopping at $s_\text{stop}$ is ultimately possible.

Due to the rational driver assumption, the road users approaching the junction from the right are not permitted to drive on the same lane as the CAV and thus are ignored.
Furthermore, the position  $s_\text{PGA}$ is extracted from $\mathcal{M}$, at which the CAV has successfully merged and thus does not have to give way any longer.
For the application scenario, all PGAs share this position $s_\text{PGA}$, since the changed traffic rules in this point mark an $\hat{\mathcal{S}}_\text{ ISS}$ start at this position.
Furthermore, the rule change marks the end $\overline{s}$ of the respective situation context.
In turn, the beginning $\underline{s}$ of the situation context is determined as the first position from which $\overline{s}$ can be reached within $T_\text{pred}$.
For the merging scenario, we use the following heuristics to assign an importance to the behavior options. 
The highest priority $I=2$ is assigned to behavior options that correspond to dynamically merging into the junction.
This is indicated by $\boldsymbol{s}_{f,N_t} = s_\text{PGA}$, i.e., the trajectory ends in $s_\text{PGA}$.
Behavior options in which the CAV gently stops at $s_\text{stop}$, i.e. with $a_\shortparallel(t_{f,N_t}) = 0$, are given the second priority $I=1$, and behavior options corresponding to the fail-safe strategy are given the lowest priority $I=0$. 

Dynamically merging consists of sequences of $N_t=3$ target states $\boldsymbol{x}_{f,i}$: the first one is a PNR, the second one the curve exit point after turning, from which on the CAV can accelerate again, and finally a PGA.
Gently stopping and the fail-safe strategy only consist of one target state $\boldsymbol{x}_{f} = [s_\text{yield},0,0]^T$. 
For all $I=0$, $I=1$, and $I=2$, multiple behavior options $\mathcal{B}_k$ are sampled.
First, the PNRs and PGAs are sampled over velocity, where the positions of the PNRs are a function of velocity, while the positions of the PGAs are fixed to $s_\text{PGA}$.
Moreover, the final times $t_{f,i}$ are sampled dynamically according to the other relevant road users.

If there is another road user ahead, the behavior options are dynamically changed.
In this case, the behavior options consist of either following the vehicle ahead and stopping at the yield line, or following the vehicle ahead to the PGA, both with $N=2$.
In each of these cases, $v_\text{max}(s_\shortparallel)$ is modified to $v_\text{max}(s_\shortparallel,t)$ using \eqref{eq:dyn_speed_limit_profile}.
Furthermore, the target sequence
\begin{equation}
	 \langle ( [s_\text{pred}(t_{f1}),v_\text{pred}(t_{f1}),0] ,t_{f1} ), \, ( [s_{f2},v_{f2},0] ,t_{f2} ) \rangle
\end{equation}
of each behavior option $\mathcal{B}_k$ results from sampling over $t_{f,1}$ and $t_{f,2}$. 

In summary, $[{s}_j,{s}_{j+1})$ is determined from the PGA position and the first position from which it can be reached.
The set of constraints $\mathcal{S}_C$ contains $T_\text{pred}$, the dynamic constraints $a_\text{min}$ and $a_\text{max}$, and the speed limit profile $v_\text{max}(s,t)$ for the CAV.
The set of regularizing assumptions  $\mathcal{S}_\text{RA}$ contains the lanes (extracted from $\mathcal{M}$) to look for relevant road users, the maximum speed $v_\text{sl}$ for the other vehicles, the parameters of the risk model, the PNRs and PGAs, and the probability of reliable perception $p_{\text{rel}}$ determined online by the reliability estimation.
Finally, the list of behavior options $\mathcal{L}_\mathcal{B}$ contains different variations of the behavior classes of either dynamically merging into the junction, gently stopping, or fail-safe stopping as described above, all with the same $\mathcal{P}$, and is the input for the sampling scheme of the motion planning.

\section{MOTION PLANNING} \label{sec:Planning}
The motion planning scheme has three main goals while moving the CAV from its current state $\boldsymbol{x}_0$ to its final state: safety, robustness, and passenger comfort.
These goals are combined within a performance function using weighting parameters for the mathematical formulations of each goal. This performance function is then optimized with respect to several constraints to yield the optimal motion decision including the corresponding trajectory.
Besides these optimization goals, the motion planning should account for several behavior options, provide adequate timing to chosen behavior, and involve the situation context in its decision making.

To address robustness, the longitudinal and lateral planning of the trajectory are decoupled through path-velocity decomposition, which is a common strategy \cite{Hubmann2018a}.
Thus, the motion planning only needs to plan longitudinally, for which a simple linear motion model is sufficient, while the lateral trajectory tracking takes care of keeping the vehicle on the road.
Consequently, the motion planning gets more robust, as disturbances only influence the linear model, while the non-linearities are taken care off by the subsidiary control, i.e., the lateral trajectory tracking.

Since the longitudinal planning accounts for the curvature of the road, traffic rules, obstacles ahead, and the physical limits in terms of constraints, the trajectory resulting from the lateral trajectory tracking is always feasible.

To increase the robustness against perception errors, the objects are projected to the lane and treated in 1D assuming that the road users follow their lane.
By treating interactions with other road users as constraints, prediction and planning are almost completely decoupled.
Thus, the influence of possible distortions is limited and robustness is increased.
Furthermore, the decoupling allows for keeping prediction and planning in separate modules, which is particularly important if the prediction is made externally on a MEC server, like in our case.

\subsection{Integration of an External Environment Model} \label{sec:ResolvingOcclusions}
Occlusions are a frequent issue in urban areas, even among human drivers.
Due to reduced sight, road users have to slow down and have a careful look into the occluded areas before they can proceed.
Apart from low traffic efficiency, this poses high demands on the sensor setup and thus raises cost for automated vehicles.
In contrast, with external information communicated to the CAV via V2X communication, such occlusions can be resolved even before the vehicle arrives at the critical point.

\begin{figure}[tbp]
	\centering
	\includegraphics[width=0.45\textwidth]{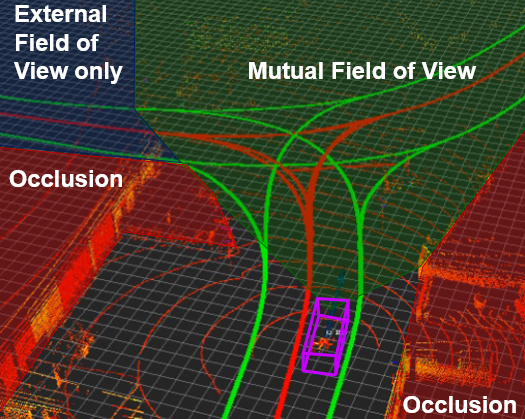}
	\caption{Mutual FOV and FOV of the external perception. Due to the occlusion, only the external perception covers the regions relevant for planning the merging maneuver.}
	\label{fig:NoFusion}
\end{figure}

In classical approaches, T2T fusion is used to combine the external and the ego environment model, as it shows benefits in mutually observed areas.
If an object in the mutual FOV is included in both environment models, its precision is increased by the fusion.
In contrast, for objects from the non-overlapping FOV, T2T essentially yields the concatenation of the ego and the external object list $\mathcal{L}_{\text{ego}}$ and $\mathcal{L}_{\text{ext}}$, respectively.
To underline this, Figure \ref{fig:NoFusion} shows an example from our application scenario.
As can be seen, there is only a limited mutual FOV which is well covered by the ego perception with sufficient accuracy.
Hence, the gain in precision for objects in this mutual FOV will hardly improve the motion planning and, thus, from an application perspective, T2T fusion is not worth the effort in this situation.

Therefore, we propose to use the external and the ego environment model in parallel instead of merging them with T2T fusion.
First, $\mathcal{L}_{\text{ego}}$ and $\mathcal{L}_{\text{ext}}$ are checked for consistency by the reliability estimation from \cite{MuellerIV2019}. If significant discrepancies occur, this indicates a malfunction of the perception and thus makes the CAV resort to relying on the more trusted $\mathcal{L}_{\text{ego}}$. Otherwise, with a simple nearest neighbor association, clearly redundant objects are identified and only the more trusted one is added to the list of relevant objects $\mathcal{L}_{\text{rel}}$. Furthermore, the situation context is used to decide which objects are relevant to the planning and thus are included in $\mathcal{L}_{\text{rel}}$, and which objects can be ignored, e.g. distant pedestrians on a sidewalk.
Based on the list of relevant objects, the motion planning then decides on a specific behavior option and the corresponding timing.

\subsection{Risk Model} \label{sec:RiskModel}

We propose a risk model that combines the set-based methods with a probabilistic approach.
With that, we obtain far-reaching safety guarantees, where we are still able to interface with probabilistic perception modules.
Furthermore, we obtain an extended solution set, since our requirements are less strict than those of the set-based methods.	
As discussed in Section~\ref{sec:contextAwareness}, the safety of a trajectory is formally guaranteed if it is fully contained withing the ISS.
Note, however, that even if a $\mathcal{T}_\shortparallel$ is not fully included in the ISS, it does not mean that $\mathcal{T}_\shortparallel$ causes a collision. It rather tells that safety is not guaranteed.
Hence, we
extend this concept by a probabilistic risk model which essentially estimates the probability $p_{\text{risk}}$ that the planned trajectory is not fully contained in the ISS.
This can occur for two reasons: first, a regularizing assumption may not hold, or second, the candidate $\mathcal{T}_\shortparallel$ is not entirely included in an $\hat{\mathcal{S}}_\text{ISS}$, so there is a risk that it is not included in the ISS either.
Hence, for $p_{\text{risk}}\neq0$, the safety guarantee is reduced to a guarantee in the probabilistic sense with the residual risk limited to an arbitrary maximum accepted residual risk $ p_{\text{risk,max}}$.

If the planned trajectory is not entirely included within the current safe set of the CAV, the concept of PNR and PGA from \cite{Pek2017} is adopted to determine the residual risk $p_{\text{risk}}$.
The residual risk is calculated as the probability that the states on the trajectory between PNR and PGA, which is called the \textit{safety critical passageway} \cite{Pek2017}, actually are not fully included in the ISS. Mathematically, this is expressed as
\begin{equation}
\small
\label{eq:RiskModel}
p_{\text{risk}}(\mathcal{T}_\shortparallel) =  (1- p_{\text{rel}}) + p_{\text{rel}} \int_{t_\text{PNR}}^{t_\text{PGA}} \!\!\! \int_{s_\shortparallel(t)-s^{-}(t)}^{s_\shortparallel(t)+s^{+}(t)}  f_{\mathcal{L}_\text{rel}\cup \mathcal{O}_\text{virt}}(s,t) \text{d}s \, \text{d}t \, .
\end{equation}
If the perception module is not reliable,
the regularizing assumptions of reliable perception breaks down and renders the calculated ISS invalid.
Thus, $(1- p_{\text{rel}})$ directly adds to $p_{\text{risk}}$.
To estimate $p_{\text{rel}}$, in our previous work \cite{MuellerIV2019}, a reliability estimation scheme was proposed that tests the incoming data for plausibility and consistency. Thus, the CAV is able to online estimate the reliability of the external information source while approaching the junction.
The reliability estimation yields a $\beta$-distribution over the probability that the information source actually is correct.
Hence, the integral from a given confidence level $\alpha$ up to $100\%$ yields
\begin{equation}
p_{\text{rel}} = \int_{\alpha}^{1} \beta(x) \text{d}x \, .
\end{equation}
In \eqref{eq:RiskModel}, $f_{\mathcal{L}_\text{rel}\cup \mathcal{O}_\text{virt}}(s,t)$ represents the joint distribution of the relevant objects and their predictions in the environment model $\mathcal{L}_\text{rel}$ and the virtual objects in $\mathcal{O}_\text{virt}$.
Virtual objects $o_j \in \mathcal{O}_\text{virt}$ are used to model epistemic uncertainty, e.g., due to remaining occlusions or the end of sight (EOS).
For the application scenario, at the EOS, a virtual vehicle approaching the junction with the velocity of the speed limit $v=v_\text{sl}$ can be added in the absence of other road users to make the CAV prefer merging early, where it is certain that no other road user interferes.

The general formula \eqref{eq:RiskModel} can be approximated for many practical cases, like our exemplary application scenario.
In that, the relevant road users are the ones on the ego lane and, while the CAV passes the yield line and merges onto its target lane (i.e., $t_\text{yield} \in [t_\text{PNR},t_\text{PGA}]$), the road users on that lane.
As the occlusions on the junction are resolved by the external information, a single virtual object at EOS is added.
All relevant road users and the CAV are projected to the target lane and the other road users' states are modeled as Gaussian distributions.
Hence, the space integral in \eqref{eq:RiskModel} can be expressed in terms of the probabilities of each object $o_{i} \in \mathcal{L}_\text{rel} \cup \mathcal{O}_\text{virt}$
\begin{equation}
\tilde{\tilde{p}}_{\text{risk},i}(\mathcal{T}_\shortparallel) = \text{erf}\bigg( \tfrac{s_\shortparallel(t)-s^{-}(t) - \mu_{o_{i}}(t) }{ \sigma_{o_{i}}(t) } \bigg)
-  \text{erf}\bigg( \tfrac{s_\shortparallel(t)+s^{+}(t) - \mu_{o_{i}}(t) }{ \sigma_{o_{i}}(t) } \bigg)
\end{equation}
that
it violates the safety distances at time $t$, where $\text{erf}(\, \cdot \,)$ is the error function of the standard normal distribution and $\mu_{o_i}$ and $\sigma_{o_i}$ describe the mean and standard deviation of the object's predicted position at time $t$.
Accumulating the $\tilde{\tilde{p}}_{\text{risk},i}(t)$ for all objects $i=1,\dots,N_o$ and all instances of time $t \in [t_\text{PNR},t_\text{PGA}]$ then yields the integral term of \eqref{eq:RiskModel}.
However, simple addition of all $\tilde{\tilde{p}}_{\text{risk},i}(\mathcal{T})$ does not yield correct results, because modeling the other road users as independent Gaussian distributions neither reflects their interdependence, nor does it account for the temporal correlation of their states.
To account for these dependencies, we model \eqref{eq:RiskModel} through the iteration
\begin{subequations}
	\label{eq:RiskModelSimplified}
	\begin{align}
	\small
	\tilde{p}_{\text{risk},i}(\mathcal{T}_\shortparallel) &= \max_{t} \{ \tilde{\tilde{p}}_{\text{risk},i}(t) \} , \; t \in [t_\text{PNR},t_\text{PGA}] \, ,\\
	p_{\text{risk},0}(\mathcal{T}_\shortparallel) &= \tilde{p}_{\text{risk},0}(\mathcal{T}_\shortparallel) \\
	p_{\text{risk},i+1}(\mathcal{T}_\shortparallel) &= p_{\text{risk},i}(\mathcal{T}_\shortparallel) + (1-p_{\text{risk},i}(\mathcal{T}_\shortparallel)) \cdot	 \tilde{p}_{\text{risk},i+1}(\mathcal{T}_\shortparallel) \, ,\\
	p_{\text{risk}}(\mathcal{T}_\shortparallel)&=  (1- p_{\text{rel}}) + p_{\text{rel}} \cdot p_{\text{risk},N_o}(\mathcal{T}_\shortparallel) \, .
	\end{align}
\end{subequations}
The iteration yields the probability that either one of the objects at its closest approaching violates the safety distances or, due to unreliable perception, the safety critical passageway is not included in a safe set.

\subsection{Holistic Behavior and Longitudinal Planning} \label{sec:ModelingOCP}
Our sampling-based approach uses an analytical solution for a simplified OCP to generate candidate longitudinal trajectories $\mathcal{T}_\shortparallel$ from samples $\mathcal{B}_k$ of the sampling space $\mathcal{L}_\mathcal{B}$ .
To obtain the trajectory candidates, the current longitudinal ego state $\boldsymbol{x}_{\shortparallel,0}$ is used as the initial state. 
Then, the longitudinal trajectory candidate $\mathcal{T}_\shortparallel$ is planned segment-wise along the target sequence $\langle(\boldsymbol{x}_f,t_f)\rangle_{i}$ of the respective behavior option $\mathcal{B}_k$.
Concatenating the segments yields $\mathcal{T}_\shortparallel$.
These candidates then are checked to see if they satisfy the full set of constraints $\mathcal{S}_C$. 
The constraints can either be expressed through $a_\text{min}$, $a_\text{max}$, and a velocity profile $v_\text{max}(s,t)$, or are reflected in the risk model, i.e., in $p_\text{risk}(\mathcal{T}_\shortparallel)$.
Finally, the overall cost of the candidates are calculated and the $\mathcal{T}_\shortparallel$ with lowest cost that fulfills all constraints is selected as the optimal trajectory $\mathcal{T}_\shortparallel^\ast$.
The $\mathcal{T}_\shortparallel$  are generated from $\mathcal{B}_k$ using the simplified OCP 
\begin{subequations}
	\label{eq:problemLong}
	\begin{align}
		\small
		u^\ast(t) &= \arg \min_{u(t)} \left \{ \int_{0}^{\Delta t_{f,i}} \tfrac{w_t+t}{2+2t} u(t)^2 \text{d}t \right \} \text{ subject to} \\
		\dot{\boldsymbol{x}}_{\shortparallel}(t) &= \left[ \begin{smallmatrix}
		0 & 1 & 0 \\
		0 & 0 & 1 \\
		0 & 0 & 0
		\end{smallmatrix} \right] \, \boldsymbol{x}_{\shortparallel}(t) + \left[ \begin{smallmatrix}
		0 \\
		0 \\
		1
		\end{smallmatrix} \right] u(t), \\
		\boldsymbol{x}_\shortparallel(0) &= \boldsymbol{x}_{\shortparallel,0} \, , \quad \boldsymbol{x}_\shortparallel(\Delta t_f) = \boldsymbol{x}_{\shortparallel,f} \, ,		
	\end{align}
\end{subequations}
where $\Delta t_{f,i} = t_{f,i}-t_{f,i-1}, \, t_{f,0}=0$ is the time difference between the current and the previous target state.
The Lagrangian term of the cost function is a function of the time weight $w_t$ and the longitudinal jerk $u(t)$, since, as discussed in our previous work \cite{MuellerITSC2019}, jerk is a common criterion for passenger comfort and time-weighting the jerk can mitigate jerky behavior due to replanning.
In \cite{MuellerITSC2019}, we have also derived the analytic solution to \eqref{eq:problemLong} and evaluated the effect of $w_t$.
Thus, in this work, for simplicity, we choose $w_t=1$.

With the risk model, $p_\text{risk}(\mathcal{T}_\shortparallel)$ is determined and $\mathcal{T}_\shortparallel$ is checked to fulfill the set of constraints $\mathcal{S}_C$.
Furthermore, the overall cost of the candidate trajectory
\begin{align}
	J_\shortparallel(\mathcal{T}_\shortparallel) &= \sum_{i=1}^{N_t}
	\int_{0}^{\Delta t_{f,i}} \tfrac{w_t+t}{2+2t}u^2(t) \text{d}t + w_{t_f} \Delta t_{f,i}^2 \nonumber \\
	& \quad + p_\text{risk}(\mathcal{T}_\shortparallel)
\end{align}
is calculated and the overall cheapest candidate is selected.
In sum, the combined behavior and longitudinal trajectory planning solves the OCP
\begin{subequations}
	\begin{align}
		\small
		u^\ast(t) &= \arg \min_{u(t)} \{ J_\shortparallel(\mathcal{T}_\shortparallel) \} \quad \text{subject to} \\
		\dot{\boldsymbol{x}}_{\shortparallel}(t) &= \left[ \begin{smallmatrix}
		0 & 1 & 0 \\
		0 & 0 & 1 \\
		0 & 0 & 0
		\end{smallmatrix} \right] \, \boldsymbol{x}_{\shortparallel}(t) + \left[ \begin{smallmatrix}
		0 \\
		0 \\
		1
		\end{smallmatrix} \right] u(t), \\
		a_\text{min} &\leq a_\shortparallel(t) \leq a_\text{max} \, , \nonumber \\
		0 &\leq v_\shortparallel(t) \leq v_\text{max}(s_\shortparallel,t) \, , \label{eq:dynamicConstraints}\\
	    0 &\leq  p_{\text{risk}}( \mathcal{T}_\shortparallel) \leq p_{\text{risk,max}} \, , \label{eq:risk1}\\	
	    \boldsymbol{x}_\shortparallel(0) &= \boldsymbol{x}_{\shortparallel,0}, \quad
	    \mathcal{T}_\shortparallel(t_{f,i}) = \boldsymbol{x}_{f,i}, \quad i = 1,\dots N_t ,		
	\end{align}
\end{subequations}
where the constraints are reflected in terms of the velocity profile $v_\text{max}(s,t)$ and the acceleration limits $a_\text{min}$ and $a_\text{max}$ from $\mathcal{S}_C$. However, other constraints could also be accounted for.

\subsection{Planning Algorithm} \label{sec:PlanningAlgorithm}
The motion planning scheme, Algorithm \ref{alg:PlanningScheme}, takes the local map $\mathcal{M}$, the current ego state $\boldsymbol{x}_0$ (as defined in \cite{Gritschneder2018}), the route $\mathcal{R}$, the ego object list $\mathcal{L}_{\text{ego}}$ from the CAV's environment model, the external object list $\mathcal{L}_{\text{ext}}$, and the list of pre-computed situation contexts $\mathcal{L}_C$ as inputs.
The object lists already contain the predictions of the respective objects for the next $T_\text{pred}$ seconds.
For initialization (see Algorithm~\ref{alg:PlanningScheme} lines~\ref{alg:InitStart}-\ref{alg:InitEnd}), the minimal trajectory cost $c_{\text{min}}$ is set to infinity and the optimal longitudinal trajectory $\mathcal{T}_\shortparallel^\ast$ is initialized with an empty set.
Furthermore, the current situation context $\mathcal{C}$ is updated as described in Section \ref{sec:contextAwareness}.
With $\mathcal{C}$, the relevant objects for planning are selected from $\mathcal{L}_{\text{ego}}$ and $\mathcal{L}_{\text{ext}}$.
Particularly, as compared to $\mathcal{L}_{\text{ego}} \cup \mathcal{L}_{\text{ext}}$, duplicates and objects on lanes unrelated to the intended motion of the CAV are not included in $\mathcal{L}_{\text{rel}}$.
Furthermore, the risk model is parametrized according to $\mathcal{C}$.

\begin{figure}
	\vspace{-2.7mm}
	\makebox[\linewidth]{
		\begin{minipage}{\dimexpr0.8\linewidth+5em}
		\begin{algorithm}[H]
			\caption{Motion Planning Scheme}
			\label{alg:PlanningScheme}
			\small
			\begin{algorithmic}[1]
				\renewcommand{\algorithmicrequire}{\textbf{Input:}}
				\renewcommand{\algorithmicensure}{\textbf{Output:}}				
				\REQUIRE Local map $\mathcal{M}$, current ego state $\boldsymbol{x}_0$, route $\mathcal{R}$, \\ $\quad \,$ ego and external object list with prediction $\mathcal{L}_{\text{ego}}$, $\mathcal{L}_{\text{ext}}$, \\ $\quad \,$ list of pre-computed situation contexts $\mathcal{L}_C$
				\ENSURE  Optimal Trajectory $\mathcal{T}^{\ast}$
				\STATE $c_{\text{min}} \gets \infty$; $\mathcal{T}_\shortparallel^{\ast} \gets \emptyset ; \quad \mathcal{B}^\ast \gets \emptyset$	\label{alg:InitStart} 		
				\STATE $\mathcal{C},\boldsymbol{x}_{\shortparallel,0} \gets$ SelectAndUpdateContext$(\mathcal{L}_{\text{ego}}, \mathcal{L}_{\text{ext}},\boldsymbol{x}_0, \mathcal{L}_\mathcal{C})$ 
				\STATE $\mathcal{L}_{\text{rel}} \gets$ ExtractObjectsForRiskModel$(\mathcal{C},\mathcal{L}_{\text{ego}}, \mathcal{L}_{\text{ext}})$ \label{alg:InitEnd}
				\FORALL {$\mathcal{B}_k \in \mathcal{L}_\mathcal{B}$} \label{alg:LongPlanStart} 
				\STATE $j, \, \Delta t_{f,j+1} \gets $ GetTargetSubsequence$(\langle(\boldsymbol{x}_f,t_f)\rangle_{i}, \, \boldsymbol{x}_{\shortparallel,0})$
					\STATE $\tilde{\mathcal{T}}_\shortparallel \gets$ AppendTrajectorySegment$(\emptyset,\boldsymbol{x}_{\shortparallel,0},\boldsymbol{x}_{f,j+1},\Delta t_{f,j+1})$
					\FOR {$i = j+1$ \TO $N_t - 1$}
					\STATE $\tilde{\mathcal{T}}_\shortparallel \gets$ AppendTrajectorySegment$(\tilde{\mathcal{T}}_\shortparallel,\boldsymbol{x}_{f,i},\boldsymbol{x}_{f,i+1},\Delta  t_{f,i+1})$		
					\ENDFOR			
					\STATE $p_\text{risk} \gets$ EvaluateRisk$(\mathcal{T}_\shortparallel,p_\text{rel},\mathcal{L}_{\text{rel}})$
					\IF {constraintCheck$(\tilde{\mathcal{T}}_\shortparallel)$ \AND $J_\shortparallel(\tilde{\mathcal{T}}_\shortparallel) < c_\text{min}$ }
					\STATE $c_\text{min} \gets J_\shortparallel(\tilde{\mathcal{T}}_\shortparallel); \quad \mathcal{T}_\shortparallel^\ast \gets \tilde{\mathcal{T}}_\shortparallel; \quad \mathcal{B}^\ast \gets \mathcal{B}_k$
					\ELSE
					\STATE \textit{continue with next behavior option} $\mathcal{B}_{k+1}$
					\ENDIF			
				\IF { $\mathcal{T}_\shortparallel^\ast \neq \emptyset$ \AND $I_{k+1} < I_k$}
				\STATE \textit{break for loop}
				\ENDIF			
				\ENDFOR \label{alg:LongPlanEnd}
				\STATE $\mathcal{T}^{\ast} \gets$ LateralTrajectoryTracking$(\mathcal{T}^{\ast}_\shortparallel,\mathcal{P}_{\mathcal{B}^\ast},\boldsymbol{x}_0)$
				\RETURN  $\mathcal{T}^{\ast}$
			\end{algorithmic}
		\end{algorithm}
	\end{minipage}}
\end{figure}

Next, the algorithm explores the behavior options $\mathcal{B}_k$ available in $\mathcal{L}_\mathcal{B}$ and determines the optimal longitudinal trajectory $\mathcal{T}_\shortparallel^\ast$ as described in Section~\ref{sec:ModelingOCP}. This is reflected in Algorithm~\ref{alg:PlanningScheme} lines~\ref{alg:LongPlanStart}-\ref{alg:LongPlanEnd}.
If one priority class is completely explored and a valid solution is found, i.e. $I_{k+1} < I_k$ and $\mathcal{T}_\shortparallel^\ast \neq \emptyset$, the search is terminated in advance.
Otherwise, the search goes on until finally one of the fail-safe options is chosen, which is always contained in $\mathcal{L}_\mathcal{B}$.

Lastly, the optimal longitudinal trajectory $\mathcal{T}^{\ast}_\shortparallel$ is fed into the lateral trajectory tracking to calculate the full, drivable trajectory $\mathcal{T}^{ \ast}$.
The trajectory is optimized numerically using the optimization tool GRAMPC \cite{Englert2019} and a non-linear bicycle model \cite{Gritschneder2018}.
To do so, we follow the non-linear model predictive control (N-MPC) approach proposed by Gritschneder et al. \cite{Gritschneder2018}.
Yet, in contrast to \cite{Gritschneder2018}, in this work, the longitudinal reference trajectory and, thus, the timing of the planned maneuver, is much more important and more specific.
Therefore, while Gritschneder et al. focus on optimizing lateral quantities like the lateral velocity $v_\perp$, the lateral acceleration $a_\perp$, or the trajectory's curvature $\kappa_v$, in this work, the lateral quantities are mostly treated as constraints, optimizing only for the most relevant lateral quantities, i.e., the lateral distance from the reference $d_{\perp}$, the vehicle's orientation $\phi$, and the change rate $\dot{\delta}$ of the steering wheel angle $\delta$, where the latter is the lateral control input of the system dynamics.
Thus, the optimality criterion for the lateral trajectory tracking is altered to
\begin{align}
	J_{\text{lat}} &= \int_{0}^{t_{f,\perp}}w_v (v-v_\shortparallel)^2 + w_a (a - a_\shortparallel)^2  \nonumber \\
	&+ w_{\dot{\delta}} \dot{\delta}^2 + w_{d_{\perp}} (d_{\perp} - d_{\perp, \text{ref}})^2 + w_{\phi} (\phi-\phi_{\text{ref}})^2 \text{d}t \, ,
\end{align} where
$w_v$, $w_a$, $ w_{\dot{\delta}}$, $w_{d_{\perp}}$, and $w_{\phi}$ are the weights balancing the respective optimization goals and $t_{f,\perp}$ is the optimization horizon for the lateral trajectory tracking.
A further difference in our work is that inequality constraints are treated explicitly through the augmented Lagrangian version of GRAMPC, while Gritschneder et al. had to use a penalty function, as the augmented Lagrangian version of GRAMPC was not yet released back then.
Particularly, in this work, the lateral acceleration is treated as an inequality constraint
\begin{equation}
		a_{\perp}^2 = \left( \frac{\delta v_\shortparallel^2}{l (1 + (\tfrac{v_\shortparallel}{v_{\text{char}}})^2)}\right)^2 \leq a_{\perp,\text{max}}^2 \label{eq:LatInequalityConstraint}
\end{equation}
rather than as an optimization goal.
In \eqref{eq:LatInequalityConstraint}, $l$ refers to the wheel base of the vehicle and $v_{\text{char}}$ is the characteristic velocity of the vehicle \cite{Gritschneder2018}.

The number of behavior options in $\mathcal{B}_k \in \mathcal{L}_\mathcal{B}$ strongly varies depending on the respective situation.
In trivial situations, such as lane keeping at constant velocity, there is just one prioritized behavior option, which is feasible and optimal for the respective situation.
Due to our heuristics in $\mathcal{S}_\text{RA}$, slightly more complex situations such as driving through curves require few (about $10$) behavior options.
Such heuristics are, e.g., to search the optimal solution in the vicinity of a precomputed guess.
This reduces the sampling efforts significantly.
Highly complex traffic situations, such as dynamically merging into a yield junction, require about $100$ behavior options.
For each behavior option, the corresponding trajectory is computed by sequentially solving the OCP~\eqref{eq:problemLong}.
However, due to our analytic solution (see Section~\ref{sec:ModelingOCP}), the trajectories corresponding to the behavior options can be calculated very quickly ($\ll1~\text{ms}$).
Calculating the overall optimal trajectory in complex situations requires up to $16~\text{ms}$, from which the lateral trajectory tracking consumes about $1~\text{ms}$.
The calculations are carried out on a Intel i7-6850K processor.
For the longitudinal planning, time horizons up to $10~\text{s}$ and more are still unproblematic and the trajectory is given in its analytic form, so any discretization is possible. In contrast, the optimization horizon of the lateral trajectory tracking is restricted to $3~\text{s}$ at a discretization of $50~\text{ms}$, which yields $60$ sample points along the trajectory.

\section{EXPERIMENTAL VALIDATION} \label{sec:ExperimentalValidation}

This section presents results of the experimental validation of the proposed motion planning scheme for the the application scenario and the setup described in Section \ref{sec:ApplicationScenario}.
The motion planning scheme is validated with respect to lane keeping and maneuver timing.
Furthermore, the effects of the risk model are studied experimentally.
We refer to our previous work \cite{MuellerITSC2019} for a simulation-based statistical evaluation of decision making as well as for a simulative evaluation of trajectory generation.
A video summarizing the experiments can be found at \cite{Mueller2020}. 

\subsection{Lane Keeping Properties of the Motion Planning Scheme} \label{sec:LaneKeeping}

In the first step, we validate that the CAV adequately extracts the situation context for the junction so that the proposed lateral trajectory tracking is capable to robustly keep the CAV on the lane.
To do so, the following metric is proposed: the CAV is approximated as a bounding box with the corner points $\boldsymbol{p}_{\text{FL}}$, $\boldsymbol{p}_{\text{FR}}$, $\boldsymbol{p}_{\text{RR}}$, $\boldsymbol{p}_{\text{RL}}$ for front left, front right, rear right, and rear left, respectively.
These points are transformed to the Frenet frame with respect to the left and right boundary line  $\mathcal{B}_l,\, \mathcal{B}_r$ of the lane
\begin{align}
	&(\boldsymbol{p}_{\text{FL}}, \, \boldsymbol{p}_{\text{FR}}, \,\boldsymbol{p}_{\text{RR}}, \,\boldsymbol{p}_{\text{RL}}) \xrightarrow{\text{Frenet}(\mathcal{B}_l,\mathcal{B}_r)} \nonumber \\ &\qquad \qquad ([s,d]_{\text{FL}}^T,[s,d]_{\text{FR}}^T,[s,d]_{\text{RR}}^T,[s,d]_{\text{RL}}^T) \; .
	\label{eq:FrenetPos}
\end{align}
Then, the metric $d_{\text{Lane}}$ is caluclated as
\begin{equation}
	d_{\text{Lane}} = \min\bigg\{ -d_{\text{FL}}, d_{\text{FR}}, d_{\text{RR}}, -d_{\text{RL}} \bigg\} \, .
	\label{eq:metric}
\end{equation}
In \eqref{eq:metric}, the signs are chosen such that $d_{\text{Lane}}$ is positive whenever the CAV is on the lane.
Hence, a positive $d_{\text{Lane}}$ measures the minimum distance of the CAV to the lane boundary, whereas the CAV has left the lane for a negative $d_{\text{Lane}}$.

\begin{figure}[bthp]
	\centering
	\includegraphics{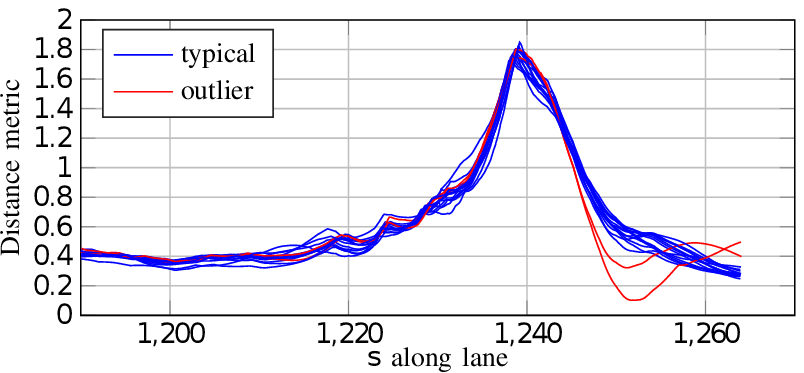}
	\caption{Evaluation of the CAV's lane keeping behavior according to the metric \eqref{eq:metric} describing the CAV's closest distance to the next lane boundary.}
	\label{fig:LaneKeepingRes}
\end{figure}

Figure \ref{fig:LaneKeepingRes} shows the evaluation of $d_{\text{Lane}}$ over $15$ automated merging maneuvers performed at the test junction.
As can be seen, the metric is always positive.
Hence, it can be concluded that the lateral trajectory tracking is able to robustly keep the CAV on track and that the situation context is adequately extracted from the map, as the vehicle arrives at the curvy junction with a velocity that the lateral trajectory tracking can handle.
The metric grows for $s \in [1220,\,1240]$, because the lane width increases within the junction and the planning tries to keep the vehicle in the middle of the lane.
For all trajectories except one, the minimum value of $d_{\text{Lane}}$ is between $20~\text{cm}$ and $40~\text{cm}$, which are typical values even for driving at straight roads.
One of the outlier trajectories comes as close as $10~\text{cm}$ to the lane boundary.
Such outliers may result from various reasons, e.g. timing issues between the end of calculation and the actuators applying the trajectory, an error in the ego state estimation, or some distorting force. Such distortion might occur due to the automatic transmission of the CAV using the clutch.
In this particular case, the evaluation of the log files indicates a drop in the accuracy of the ego state estimation.
But even then, the distance is still positive and the lane is kept.
So, in sum, Fig. \ref{fig:LaneKeepingRes} shows that the CAV robustly keeps the lane.

\subsection{Timing Behavior} \label{sec:TimingBehavior}
In the next step, we validate that the motion planning scheme reacts to oncoming traffic with reasonable behavior and that it plans appropriate longitudinal trajectories.
To do so, several highly automated test drives were conducted in real traffic, from which we report the position-over-time trajectories $s_\shortparallel(t)$.
Within these trajectories, we expect five basic merging categories: without another road user, in front of another road user, into the gap between two road users, behind another road user, and merging after stopping at the yield line.
Apart from some disturbances, which the motion controller must correct, merging without other road users is deterministic.
Hence, we expect the respective trajectories to differ only insignificantly.
Furthermore, since staying in the intersection area increases the maneuver risk, merging before another road user should typically result in trajectories very similar to those of merging without other road users.

\begin{figure}[bthp]
	\centering
	\includegraphics{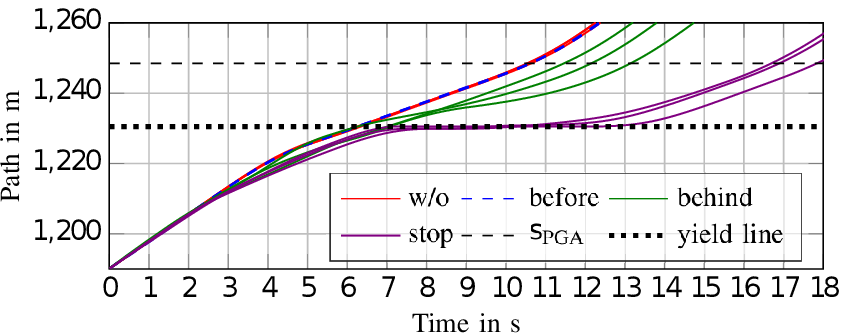}
	\caption{Trajectories for merging without other road users, before another road user, behind another road user, and merging after stopping at the yield line.}
	\label{fig:ReactionToOtherVehicles}
\end{figure}

In contrast, merging behind another road user requires the CAV to synchronize its motion to the respective road user.
Thus, the trajectories are expected to vary significantly in shape and in the arrival time at the PGA $t_f$.
Moreover, $t_f$ is expected to be higher as compared to merging without another road user.
If merging is not directly possible, the CAV stops at the yield line before merging.
Thus, the corresponding trajectories must show a horizontal segment at the yield line.
Figure \ref{fig:ReactionToOtherVehicles} inlcudes three trajectories for each of the categories except merging into a gap.
It shows that all trajectories for merging without other road users are practically identical. The same holds for merging before another road user. The minimum and the maximum arrival time $t_f$ of all these trajectories differ only by about $100~\text{ms}$.
In contrast, the trajectories of merging behind significantly vary in shape and arrival time. Here, $t_f$ varies by about $1.65~\text{s}$.
Furthermore, the trajectories of merging after stopping significantly differ in shape and arrival time. Here, $t_f$ differs by about $1.15~\text{s}$.
Trajectories of both classes show a flatter shape because the CAV slows down early, if it has to synchronize to other vehicles.
In the case that the CAV has to stop at the yield line, the trajectories even show a horizontal segment in the diagram at the position of the yield line.
This confirms the expectation and shows that the motion planning reasonably reacts to the oncoming traffic.

\begin{figure}[bthp]
	\centering
	\includegraphics{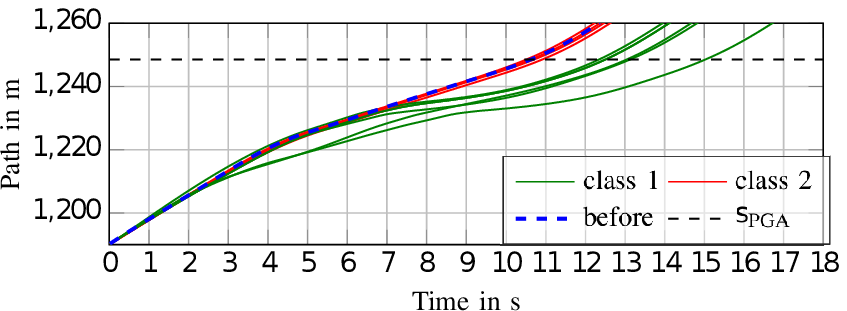}
	\caption{Timing for merging into a traffic gap. Class $1$ resembles merging behind a road user, while class $2$ behaves similar as merging before another one.}
	\label{fig:TimingGap}
\end{figure}

For merging into a traffic gap, we expect two classes of trajectories: maneuvers, for which the CAV behaves similar to merging before another road user (class 1), and maneuvers, for which the CAV behaves like merging behind another road user (class 2).
The latter occurs if the CAV needs to synchronize its motion to road user in front of the gap, while the first class results if the CAV arrives later at the yield line and mainly reacts to the road user at the end of the gap.
Figure \ref{fig:TimingGap} shows nine trajectories of the CAV automatically merging into a traffic gap between two oncoming road users.
The trajectories are plotted in red for class 1 (four trajectories) and in green for class 2 (five trajectories).
Furthermore, a trajectory for merging before another road users is plotted in dotted blue for comparison.
As can be seen, the trajectories plotted in red strongly resemble the dotted blue one, but vary a little bit more (about $0.5~\text{s}$). This can be explained by some residual effect of the road user ahead. The green trajectories vary more and are alike those of merging behind (see Fig. \ref{fig:ReactionToOtherVehicles}). Thus, these results comply with the expectations as well.

\begin{figure}[bthp]
	\centering
	\includegraphics{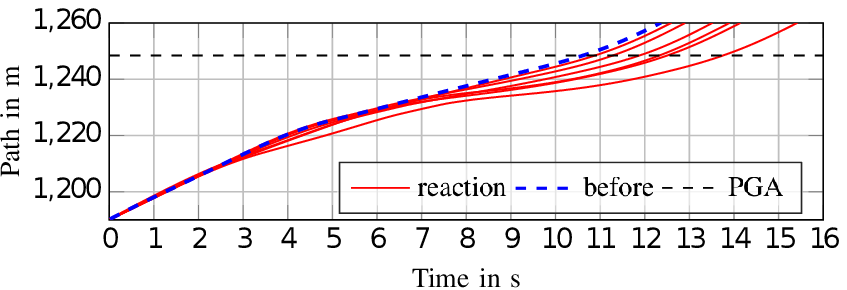}
	\caption{Trajectories corresponding to merging maneuvers with changing environment model.}
	\label{fig:TimingBefore}
\end{figure}

Finally, the ability of the motion planning scheme to react to a changing environment is evaluated.
For that, we selected scenarios where oncoming road users on the main road turn right instead of continuing on the lane to which the CAV intends to merge.
Thus, a traffic gap opens up, in which the CAV can merge.
For these scenarios, we expect strongly varying trajectories depending on when the CAV alters its maneuver decision due to the changed environment.
Particularly for tight maneuver decisions, an update in the external environment model
and the corresponding uncertainties might significantly shift the risk estimation according to the risk model and induce the motion planning to switch between different maneuver options.
Thus, the CAV might keep its velocity at some replanning cycles while it might opt to gently break in others.
Figure~\ref{fig:TimingBefore} presents $6$ trajectories showing how the CAV handles these situations.
As expected, the trajectories differ strongly in shape and arrival time $t_f$ (by about $2.75~\text{s}$). Overall, the evaluation showed that the motion planning works reliable in all different situations of the merging scenario and yields reasonable trajectories.

\subsection{Examples for Full Trajectories}\label{sec:FullTrajectories}
\begin{figure}
	\centering
	\includegraphics{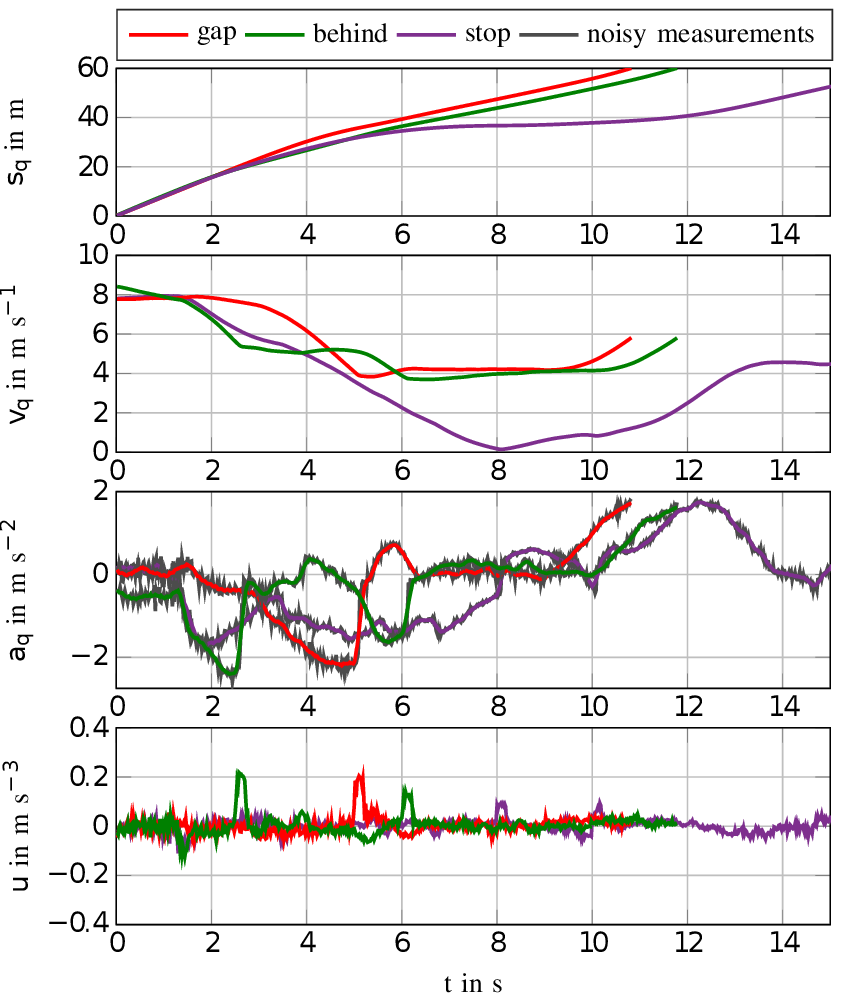}
	\caption{Example trajectories recorded during the experiments. The peaks in the jerk are due to non-compensated non-linearities in the actuation.}
	\label{fig:ExampleTrajectories}
\end{figure}

To give an impression how the actually driven trajectories look like, three representative examples are plotted in Fig.~\ref{fig:ExampleTrajectories}: one stopping trajectory (purple), one trajectory in which the CAV merges into a gap (red), and one trajectory, where the CAV merges fluently behind another road user (green).
Trajectories of merging before another road user and maneuvers without other road users are most similar to the merging trajectory. 
During the experiments, the longitudinal states $\boldsymbol{x}_\shortparallel$ have been recorded, while the longitudinal jerk had to be recovered from numerical differentiation.
To avoid noise amplification, the noisy acceleration measurements have been smoothed using a moving average filter.
The noisy original measurements have been plotted in gray, while the smoothed data are depicted in their respective color.
Except for some peaks due to uncompensated non-linearities in the actuation, the absolute longitudinal jerk always remains below $0.1~\text{m~s}^{-3}$.
Nonetheless, even at the peaks, the jerk does not exceed $0.25~\text{m~s}^{-3}$.
These are good results, since according to the study of Wang et al. \cite{Wang2020}, jerk up to about $1.5~\text{m~s}^{-3}$ is considered comfortable by passengers of automated vehicles.

\section{CONCLUSIONS AND FUTURE WORK} \label{sec:Conclusion}
In this work, a motion planning scheme for urban merging scenarios was proposed that optimizes for safety and passenger comfort over multiple behavior options.
Thus, high-level decision making and low level trajectory planning are solved holistically in one OCP, guaranteeing the feasibility of the resulting trajectory with respect to the vehicle dynamics. The parametrization of the OCP according to the dynamic situation features context awareness of our approach. The scheme is based on a parallel usage of the on-board and possibly an external environment model to resolve occlusions, where uncertainties are governed by a risk model. This risk model preserves the safety guarantees of the original set-based approach in a probablistic sense for complex scenarios.
The proposed merging planning scheme was validated experimentally on public roads with a prototype CAV, confirming that the proposed merging planning scheme works very well.

In future work, we want to evaluate the proposed merging planning scheme for further scenarios such as signalized intersections, left-turning, roundabouts, or overtaking maneuvers.
Additionally, cooperative approaches will be investigated.

\ifCLASSOPTIONcaptionsoff
  \newpage
\fi

\bibliographystyle{IEEEtran}
\bibliography{IEEEabrv,JCM_ITS_Transaction_2020.bib}

\begin{IEEEbiography}[{\includegraphics[width=0.9in,height=1.125in,clip,keepaspectratio]{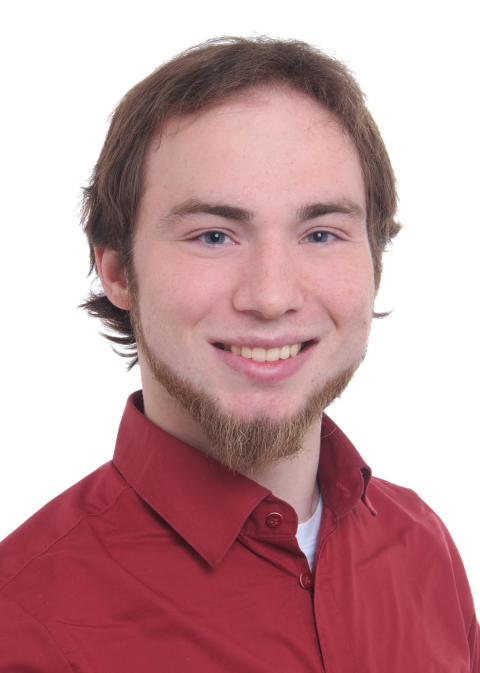}}]{Johannes Müller}
	received his bachelor’s and master’s
	degree in electrical engineering and information
	technology from Karlsruhe Institute of Technology,
	Germany in 2013 and 2016, respectively.
	He is currently a researcher at the Institute of Measurement, Control, and Microtechnology at Ulm University, working towards his PhD degree in the field of motion planning and reliability estimation for connected automated driving.
\end{IEEEbiography}	

\begin{IEEEbiography}[{\includegraphics[width=0.9in,height=1.125in,clip,keepaspectratio]{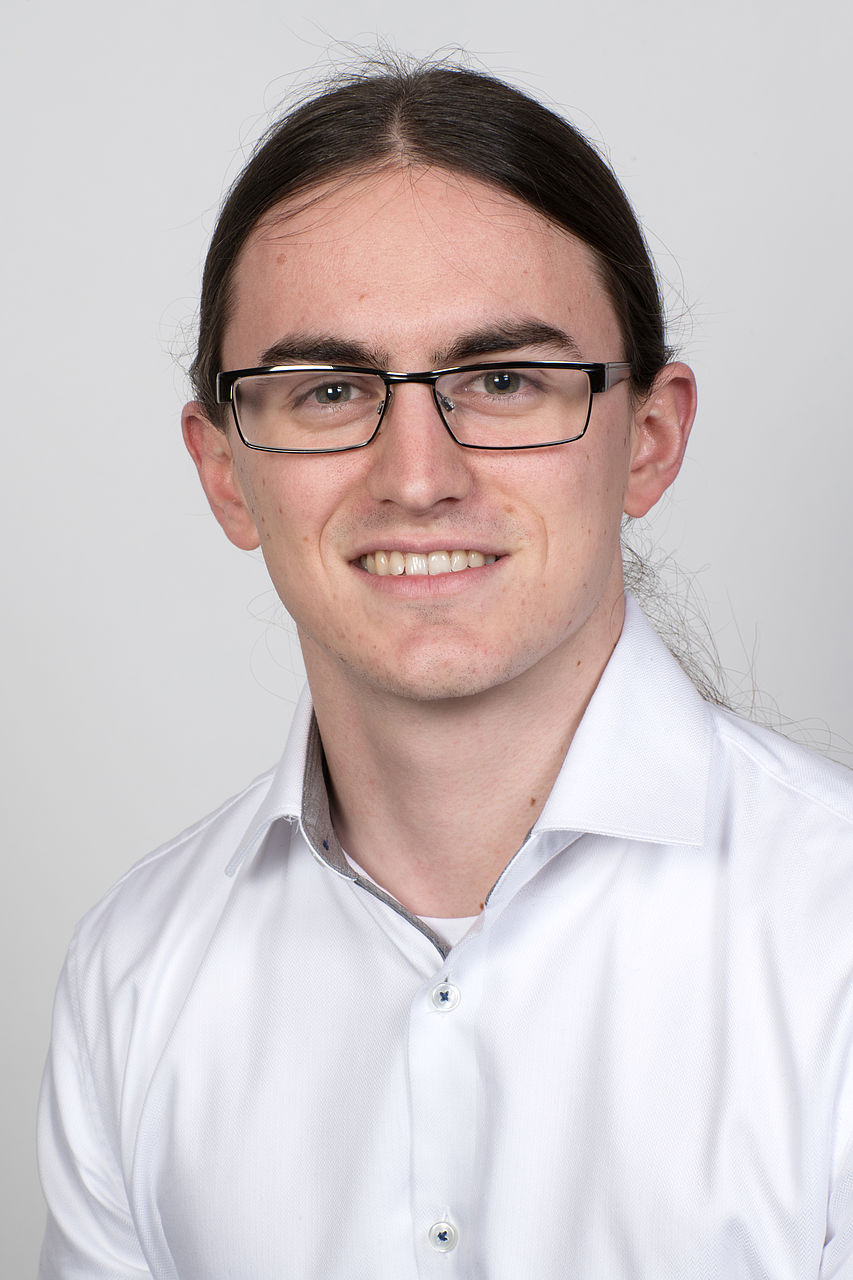}}]{Jan Strohbeck}
	received his received his Bachelor of Engineering degree in Information Technology and Master of Science degree in Computer Science from Aalen University, Germany.  Since 2018, he is working as a researcher at the Institute of Measurement, Control and Microtechnology at Ulm University, Germany.  His research focus is motion prediction of traffic participants, specifically at intersections, and the estimation of the prediction’s inherent uncertainty.
\end{IEEEbiography}	

\begin{IEEEbiography}[{\includegraphics[width=0.9in,height=1.125in,clip,keepaspectratio]{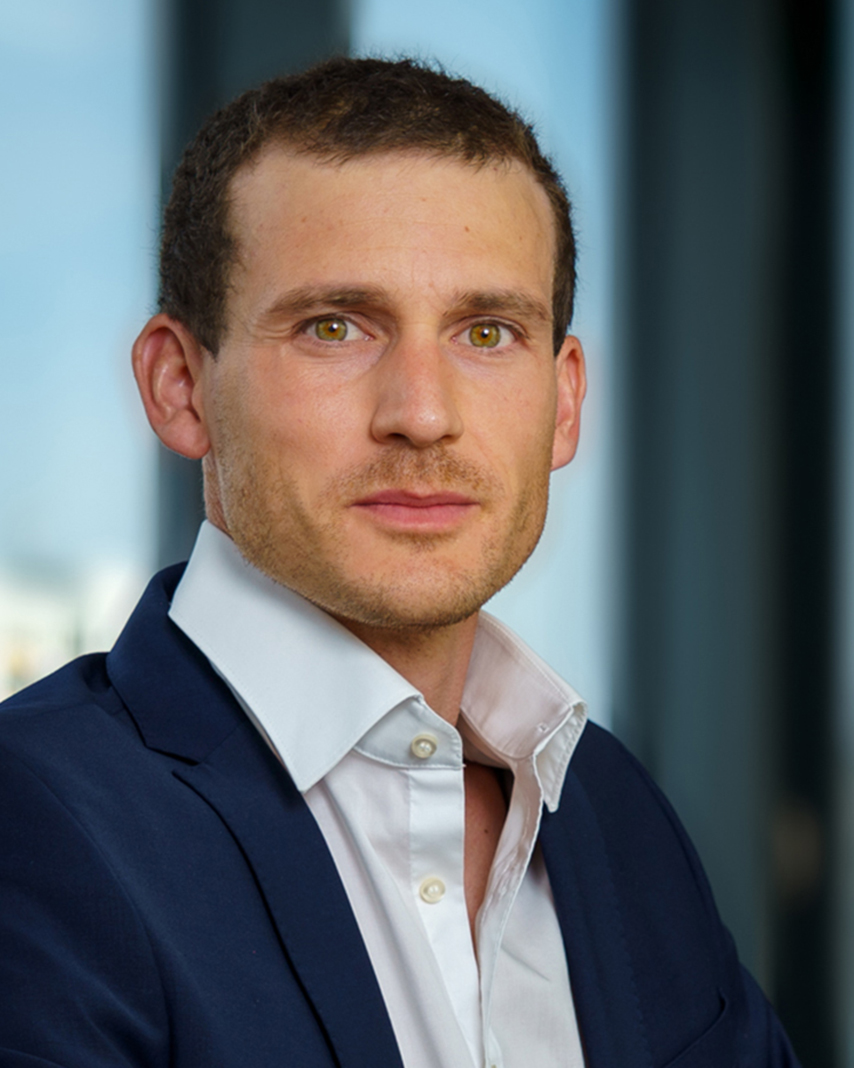}}]{Martin Herrmann}
	received his Bachelor of Science degree in Media Technology in 2014 from Ilmenau University of Technology, Germany, and his Master of Science degree in Electrical Engineering in 2017 from Ulm University, Germany. Since 2017, he is a research assistant at the Institute of Measurement, Control and Microtechnology at Ulm University, Germany. His research interests are sensor data fusion, filtering and estimation, signal processing and environment perception for autonomous driving.
\end{IEEEbiography}

\begin{IEEEbiography}[{\includegraphics[width=0.9in,height=1.125in,clip,keepaspectratio]{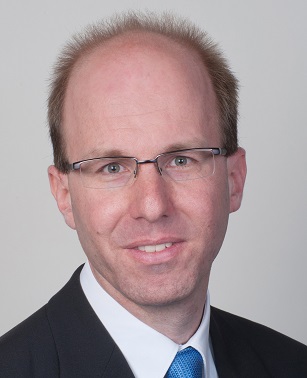}}]{Michael Buchholz}
	received his Diploma degree in Electrical Engineering and Information Technology as well as his Ph.D. from the faculty of Electrical Engineering and Information Technology at Karlsruhe Institute of Technology, Germany.  Since 2009, he is serving as a research group leader and lecturer at the Institute of Measurement, Control and Microtechnology at Ulm University, Germany.  His research interests comprise connected automated driving, electric mobility, modelling and control of mechatronic systems, and system identification.
\end{IEEEbiography}

\end{document}